\documentclass{article}
\usepackage{spconf}
\usepackage{amssymb,amsfonts}
\usepackage{amsmath,graphicx}
\usepackage{algorithmic}
\usepackage{graphicx}
\usepackage{textcomp}
\usepackage{xcolor}

\usepackage{hyperref}
\usepackage{wrapfig}
\usepackage{lscape} 
\usepackage{booktabs}
\usepackage{multirow}
\usepackage{hyperref}
\usepackage{url}
\usepackage{subcaption}
\usepackage{varwidth}
\usepackage{amsmath}
\usepackage{xspace}
\usepackage{mathptmx}
\usepackage{wrapfig}
\usepackage{enumitem}
\usepackage{hyphenat}
\usepackage{color, colortbl}
\definecolor{Gray}{gray}{0.9}

\newcommand{\ie}{\textit{i.e.}\xspace}
\newcommand{\eg}{\textit{e.g.}\xspace}
\newcommand{\xhdr}[1]{\vspace{1.3mm}\noindent{{\bf #1.}}}
\newcommand{\trs}[2]{#1_{#2}}

\title{Recursive Input and State Estimation: A General Framework for Learning from Time Series with Missing Data}
\name{Alberto García-Durán \qquad Robert West}
\address{École Polytechnique Fédérale de Lausanne (EPFL), Lausanne, Switzerland}
\begin{document}
\ninept
\maketitle
\begin{abstract}
Time series with missing data are signals encountered in important settings for machine learning. Some of the most successful prior approaches for modeling such time series are based on recurrent neural networks that transform the input and previous state to account for the missing observations, and then treat the transformed signal in a standard manner.
In this paper, we introduce a single unifying framework, \textit{Recursive Input and State Estimation} (\textsc{Rise}), for this general approach and reformulate existing models as specific instances of this framework. We then explore additional novel variations within the \textsc{Rise} framework to improve the performance of any instance. We exploit representation learning techniques to learn latent representations of the signals used by \textsc{Rise} instances. We discuss and develop various encoding techniques to learn latent signal representations. We benchmark instances of the framework with various encoding functions on three data imputation datasets, observing that \textsc{Rise} instances always benefit from encoders that learn representations for numerical values from the digits into which they can be decomposed.
\end{abstract}
\section{Introduction}
Many machine learning settings involve data consisting of time series of observations. Due to various reasons, observations may be missing from such time series. For instance, it may be impossible to observe the data during a given time window, the data-recording system may fail, or measurements may be recognized as noisy and immediately discarded at the source. Historically, the prevalent approach to handling missing data has been to apply a preprocessing step to replace the missing observations with substituted values (\eg the average of the observed values) and then treat the time series as though it were complete \cite{schafer2002missing}.
Multiple recent works \cite{lipton2016directly,yoon2017multi,yoon2018deep,che2018recurrent,cao2018brits}, however, circumvent this two-step approach and integrate a mechanism to deal with missing observations while simultaneously performing the downstream task. At their core, these approaches employ recurrent neural networks (RNN) \cite{rumelhart1986learning} whose input and hidden state are modified to account for the missing observations.

In this paper, we note fundamental commonalities between these prior approaches and define a unifying framework, \textit{Recursive Input and State Estimation} (\textsc{Rise}), that encompasses those approaches as specific instances. Instances of \textsc{Rise} operate recursively on all intermediate time steps between two observed values, by transforming input and previous state based on the preceding---and sometimes subsequent---observed values. These \textsc{Rise} instances have been proven very successful in imputation---and other downstream tasks \cite{lipton2016directly,yoon2017multi,yoon2018deep,che2018recurrent,cao2018brits}---outperforming both recurrent and non-recurrent methods such as \textsc{Mice} \cite{buuren2010mice}, matrix factorization techniques \cite{Schnabel2016RecommendationsAT} and algorithms based on expectation maximization \cite{garcia2010pattern} and expert knowledge \cite{kim2018temporal}, to name but a few. The differences across instances of the framework rely on the choice of the transformations to be applied to the input and previous state. The design of the \textsc{Rise} instances---the ``transformations''---is often tailored to a specific domain (\eg clinical data), inspired by a natural model \cite{kim2018temporal}, or guided by certain assumptions about the data. However, for a given dataset and task it is hard to foresee which instance will perform the best. 

Without loss of generality we define the transformations of the \textsc{Rise} instances to learn from univariate time series with missing data. Univariate time series are important in many machine learning problems~\cite{mhaskar2017deep,fox2018deep,papacharalampous2018predictability}. In contrast to previous works, we shift the focus from the transformations to be applied to the input and previous state, and explore encoding functions to learn latent representations for the 1D signals used by the \textsc{Rise} instances. Examples of such signals are the time gap between two observations, the value of an observation, or the average value of the time series. While there is an extended literature about learning representations for categorical data, relatively little work exists on learning representations for single continuous values. We discuss and develop appropriate encoders and empirically show that digit-based encoders improve the performance of any instance of the \textsc{Rise} framework. 

Beyond such specific improvements, however, our main contribution is more general:
whereas the earlier literature consisted of many isolated contributions, \textsc{Rise} provides a unifying framework for systematically reasoning about an entire family of techniques.


\begin{figure}[h]
\begin{subfigure}[h]{0.6\linewidth}
\includegraphics[width=\columnwidth]{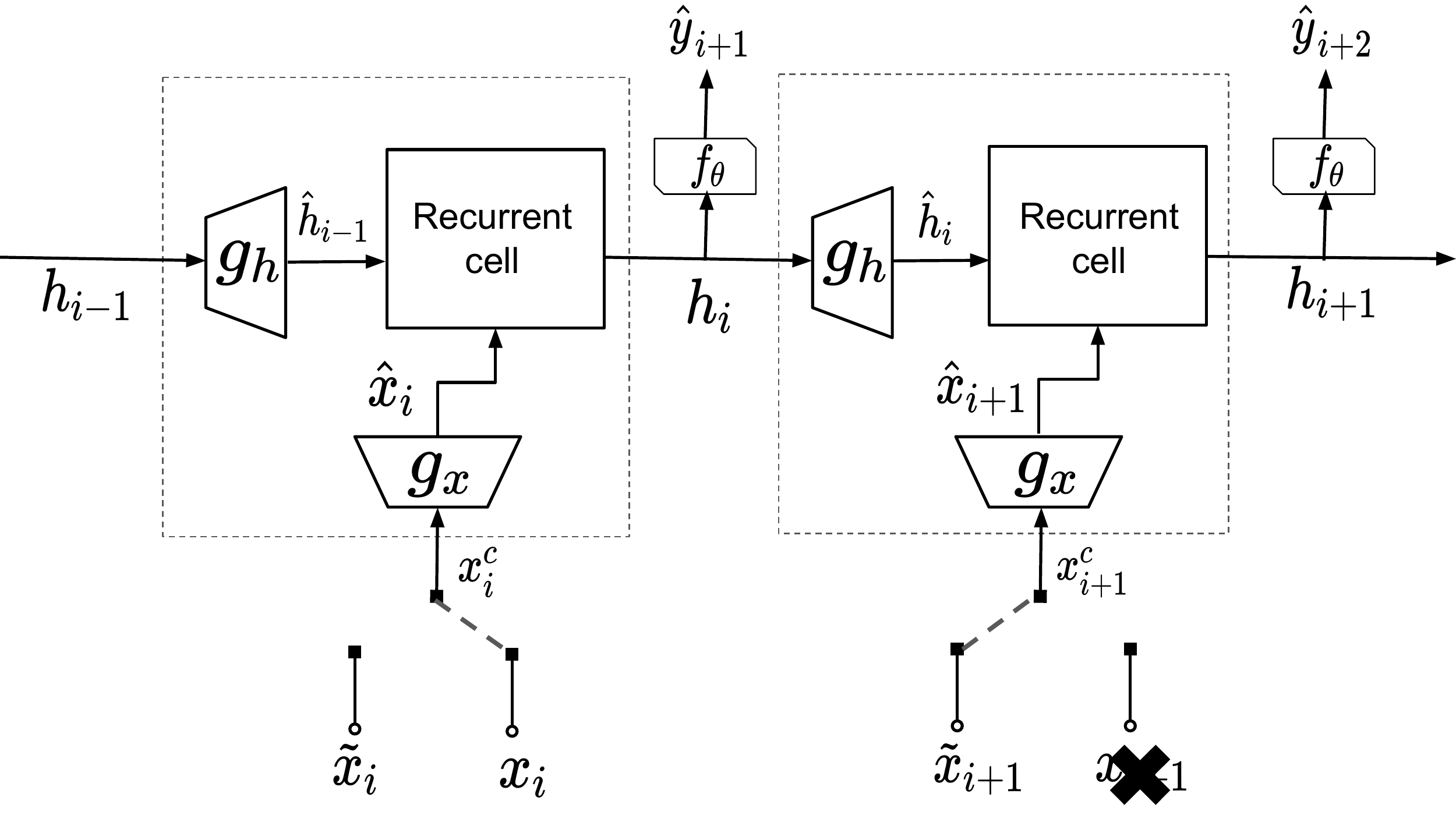}
\end{subfigure}%
\begin{subfigure}[h]{0.45\linewidth}
\includegraphics[width=\columnwidth]{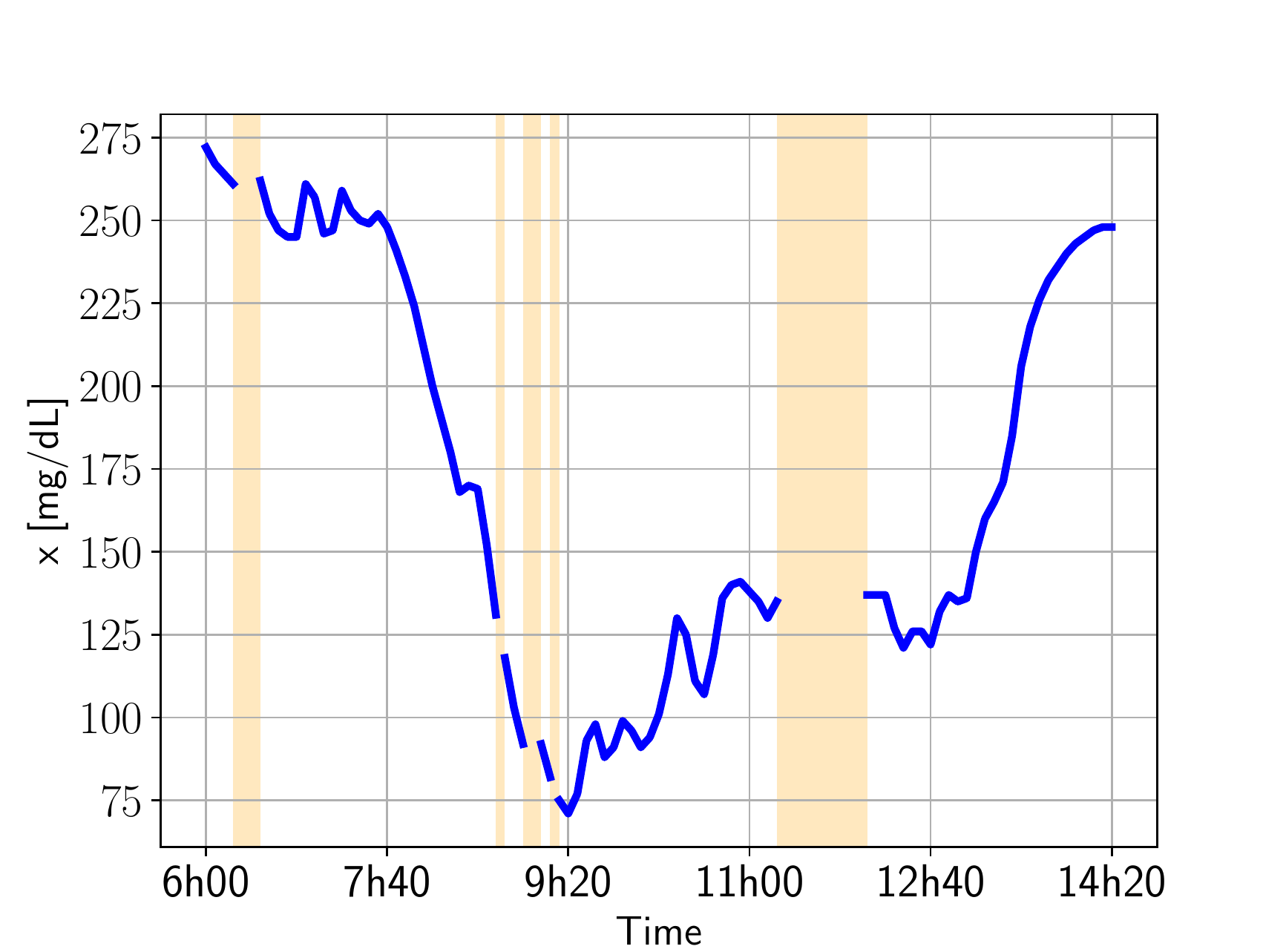}
\end{subfigure}%
\caption{\label{architecture} (Left) Architecture of the \textsc{Rise} framework. (Right) Time series with missing observations. \textsc{Rise} instances trigger the replacement input $\tilde{x}$ in those timestamps when an observation is not available (yellow background).}
\end{figure} 
\vspace{-18pt}

\begin{table*}[h!]
\centering
\caption{\label{instances:rise}  Instances of \textsc{Rise} for univariate time series. The same instances can be accommodated to the multivariate case accordingly. $\boldsymbol{\cdot}$ is the dot product. $[a;b]$ is the concatenation of terms $a$ and $b$. $\odot$ is the element-wise multiplication. $\overline{m}$ is the complement of the binary value $m$. $\trs{x}{i'}$ is the value of the last time ($i' < i$) the signal was observed. \textbf{Upper~(white):} Instances that transform input and incoming hidden state based on only preceding observed values. \textbf{Lower (gray):} Instances that leverage all observed values of the time series. $x_{av}$ is the average of the non-missing values of vector $\mathbf{x}$. Luo \textit{et al.} \cite{luo2018multivariate} is framed in a generative adversarial network, wherein a generator $G$ is fed with a Gaussian noise $z$. $\trs{\gamma}{i''}$ is a discount factor computed with respect to $\trs{x}{i''}$, which is the nearest observed value in either time direction, and $\tau$ is a threshold value. An asterisk * indicates that the described equations are computed in a bidirectional manner, and the final prediction results of the aggregation of both forward and backward predictions.}
\resizebox{0.64\linewidth}{!}{%
\begin{tabular}{|c||l|l|l|}
\hline
 Instance & \multicolumn{1}{c|}{Replacement Input $\tilde{x}$} & \multicolumn{1}{c|}{$g_x$} & \multicolumn{1}{c|}{$g_h$} \\ \hline \hline
Simple Recursion &  $\trs{\tilde{x}}{i} = \mathbf{w}_x \boldsymbol{\cdot} \trs{\mathbf{h}}{i-1} + b_x$ 
& $\trs{\hat{x}}{i} = \trs{x}{i}^c$
& $\trs{\hat{\mathbf{h}}}{i-1} =  \trs{\mathbf{h}}{i-1}$  \\ \hline
Zero-Filling\&Indicators & $\trs{\tilde{x}}{i} = 0$  
& $\trs{\hat{\mathbf{x}}}{i} = [\trs{x}{i}^c; \trs{\overline{m}}{i}]$  
& $\trs{\hat{\mathbf{h}}}{i-1} =  \trs{\mathbf{h}}{i-1}$ \\ \hline
Forward-Filling\&Indicators & $\trs{\tilde{x}}{i} = \trs{x}{i'}$  
& $\trs{\hat{\mathbf{x}}}{i} = [\trs{x}{i}^c; \trs{\overline{m}}{i}]$  
& $\trs{\hat{\mathbf{h}}}{i-1} =  \trs{\mathbf{h}}{i-1}$ \\ \hline
\textsc{Rits-I} & $\trs{\tilde{x}}{i} = \mathbf{w}_x \boldsymbol{\cdot} \trs{\mathbf{h}}{i-1} + b_x$  
& $\trs{\hat{\mathbf{x}}}{i} = [\trs{x}{i}^c; \trs{m}{i}]$  
& $\trs{\hat{\mathbf{h}}}{i-1} =  \trs{\mathbf{\gamma}}{i}^h \odot \trs{\mathbf{h}}{i-1}$ \\ \hline \hline
\rowcolor{Gray}
\textsc{Gru-D} & $\trs{\tilde{x}}{i} = \trs{\gamma}{i}^x \trs{x}{i'} + (1-\trs{\gamma}{i}^x) x_{av}$  
& $\trs{\hat{\mathbf{x}}}{i} = [\trs{x}{i}^c; \trs{m}{i}]$ 
& $\trs{\hat{\mathbf{h}}}{i-1} =  \trs{\mathbf{\gamma}}{i}^h \odot \trs{\mathbf{h}}{i-1}$ \\ \hline
\rowcolor{Gray}
Luo \textit{et al.} \cite{luo2018multivariate} & $\trs{\tilde{x}}{i} = G(z) $  
& $\trs{\hat{x}}{i} = \trs{x}{i}^c$ 
& $\trs{\hat{\mathbf{h}}}{i-1} =  \trs{\mathbf{\gamma}}{i}^h \odot \trs{\mathbf{h}}{i-1}$ \\ \hline
\rowcolor{Gray}
Kim and Chi \cite{kim2018temporal} & $\trs{\tilde{x}}{i} =  \mathbf{1}_{\trs{\gamma}{i''} > \tau} \trs{x}{i''} + (1-\mathbf{1}_{\trs{\gamma}{i''} > \tau}) x_{av} $  
& $\trs{\hat{x}}{i} = \trs{x}{i}^c$ 
& $\trs{\hat{\mathbf{h}}}{i-1} =  \trs{\mathbf{h}}{i-1}$ \\ \hline
\rowcolor{Gray}
Yoon \textit{et al.} \cite{yoon2017multi}$^*$ & $\trs{\tilde{x}}{i} = 0 $  
& $\trs{\hat{\mathbf{x}}}{i} = [\trs{x}{i}; \trs{\delta}{i}]$ 
& $\trs{\hat{\mathbf{h}}}{i-1} =  \trs{\mathbf{h}}{i-1} $ \\ \hline
\rowcolor{Gray}
\textsc{Brits}-$\text{I}^*$ & $\trs{\tilde{x}}{i} = \mathbf{w}_x \boldsymbol{\cdot} \trs{\mathbf{h}}{i-1} + b_x$    
& $\trs{\hat{\mathbf{x}}}{i} = [\trs{x}{i}^c; \trs{m}{i}]$  
& $\trs{\hat{\mathbf{h}}}{i-1} =  \trs{\mathbf{\gamma}}{i}^h \odot \trs{\mathbf{h}}{i-1}$ \\ \hline
\end{tabular}}
\vspace{-5pt}
\end{table*}

\section{\textsc{Rise} Framework}
\label{sec:rise}
\xhdr{Notation} A univariate time series \textbf{x} = [$\trs{x}{1}$, $\trs{x}{2}$, \dots, $\trs{x}{N}$] is a sequence of $N$ scalar observations. Let \textbf{t} be the time vector [$\trs{t}{1}$, $\trs{t}{2}$, \dots, $\trs{t}{N}$], each value corresponding to the time---measured as the running time elapsed since the beginning of the time series---when the respective observation was taken.  In practice, some observations may be missing, for which an $N$-dimensional masking vector \textbf{m} is defined: $\trs{m}{i} =1$ if $\trs{x}{i}$ is observed and $\trs{m}{i} =0$ otherwise. \\

\xhdr{Unifying Framework} Instances of \textsc{Rise} are designed to operate on time series with missing data, where two consecutive observed values, taken at $\trs{t}{i}$ and $\trs{t}{i+n}$, are separated by $n-1$ missing observations. They are given by recurrent neural architectures to learn a model for predicting $p(\trs{y}{i+n} | \trs{\mathbf{x}}{1:i})$, where $y$ is the target variable, by recursively predicting all intermediate conditional terms.

While one can find a number of variants of recurrent neural architectures in the literature, all variants define a cell whose (hidden) state $\trs{\mathbf{h}}{i} \in \mathbb{R}^{d_h}$---$d_h$ is the number of units---updates based on the previous state $\trs{\mathbf{h}}{i-1}$ and current input $\trs{x}{i} \in \mathbb{R}$. In time series with missing data, instances of the \textsc{Rise} framework replace the standard input $x_i$ with a transformed input $\hat{x}_i$. Similarly, at the input of the cell the previous state $\trs{\mathbf{h}}{i-1}$ is substituted with a transformed hidden state $\trs{\hat{\mathbf{h}}}{i-1}$ to account for the last time a value was observed in the time series. The hidden state $\trs{\mathbf{h}}{i}$ is updated based on these transformed signals and the equations of the chosen recurrent architecture. The transformed input $\hat{x}_i$ depends on the conditionally replaced input $x^c_i$: $\trs{x}{i}^c = \trs{x}{i} \trs{m}{i} + (1 - \trs{m}{i}) \trs{\tilde{x}}{i},$
where $\tilde{x}_i$ is the replacement input, whose computation is instance-specific. The transformed input $\hat{x}_i$ and hidden state $\hat{\mathbf{h}}_i$ are obtained by applying functions $g_x$ and $g_h$ to the conditionally replaced input $x^c_i$ and hidden state $\mathbf{h}_i$, respectively. Figure \ref{architecture} illustrates the general architecture of the framework.

It is common to some of these works to define a time gap vector $\boldsymbol{\delta}$, defined as the time gap from the last observed value to the current timestamp. More formally:

\begin{equation}
    \trs{\delta}{i} = \left\{
                \begin{array}{l l}
                  \trs{t}{i} - \trs{t}{i-1} + \trs{\delta}{i-1}, & i>1, \trs{m}{i-1} = 0\\
                  \trs{t}{i} - \trs{t}{i-1}, & i>1, \trs{m}{i-1} = 1\\
                  0, & i = 1
                \end{array}
              \right.
    \label{delta}
\end{equation}

The time gap values are used in the computation of the so-called discount factors, denoted as $\gamma^x \in \mathbb{R}$ and $\mathbf{\gamma}^h \in \mathbb{R}^{d_h}$, which in turn are used in the computation of the transformed input and the transformed hidden state, respectively. The discount factors are defined as:
\begin{align}
  \begin{split}
    \trs{\gamma}{i}^x &= \gamma^x(\trs{\delta}{i}) = \exp(-\max(0, w^x_{\gamma} \trs{\delta}{i} + b^x_{\gamma})), \\
    \trs{\mathbf{\gamma}}{i}^h &= \mathbf{\gamma}^h(\trs{\delta}{i}) = \exp(-\max(0, \mathbf{w}^h_{\gamma} \trs{\delta}{i} + \mathbf{b}^h_{\gamma})),
  \end{split}
  \label{discount}
\end{align}
where $w^x_{\gamma}$, $b^x_{\gamma}$, $b^h_{\gamma} \in \mathbb{R}$ and $\mathbf{w}^h_{\gamma} \in \mathbb{R}^{d_h}$,  are parameters trained jointly with all other parameters of the model; and $\exp$ and $\max$ are applied element-wise. The motivation of the discount factor comes from the intuition that the influence of the past history in the current moment fades away over time.

The \textsc{Rise} instances have been proven successful in various problems involving time series with missing observations. In this work we focus on predicting substituted values for the missing observations. In this task---coined data imputation---a function $f_{\mathbf{\theta}}$, parameterized with weights $\mathbf{\theta}$, is applied to the hidden state $\mathbf{h}_i$ to predict the next (probably missing) observation: $\trs{\hat{y}}{i+1} = f_{\mathbf{\theta}}(\trs{\mathbf{h}}{i})$. The loss at the $i$-th observation is defined as $\trs{l}{i} = \trs{m}{i+1} \mathcal{L}(\trs{y}{i+1}, \trs{\hat{y}}{i+1})$, and the total loss $l$ is computed as $l = \sum_{i=1}^{N-1} l_i$. Typically, the function $f_{\theta}$ is a regression function, $\mathcal{L}$ is the mean squared error loss, and $\trs{y}{i}$ amounts to $\trs{x}{i}$---\ie the $i$-th value of the time series. Alternatively, it can be cast as a multi-class classification problem \cite{Oord2016WaveNetAG,fox2018deep}, wherein the function $f_{\theta}$ outputs a probability mass function, $\mathcal{L}$ is the cross-entropy loss, and $\trs{y}{i}$ is the one-hot encoding of the value $\trs{x}{i}$.

\xhdr{\textsc{Rise} instances from the literature}
A non-exhaustive list of recent instances of the \textsc{Rise} framework published in top-tier machine learning conferences is given in Table~\ref{instances:rise}. This list includes: simple recursion \cite{fox2018deep}, forward/zero-filling\&indicators \cite{lipton2016directly}, \textsc{Rits-I} \cite{cao2018brits}, and \textsc{Gru-D} \cite{che2018recurrent}, among others.

\section{Learning Representations for \textsc{Rise}}
\label{sec:repr}
Guided by expert knowledge or intuition, previous works have explored different mechanisms to transform the standard input and hidden state of the recurrent neural architecture. For example, \textsc{Gru-D} applies an exponential decay mechanism to the input so as to mimic clinical data, wherein the influence of the observations are expected to fade away over time. Consequently, performance across methods may vary from application to application, or even from dataset to dataset. We leverage the strength of the \textsc{Rise} framework and, different to previous instances, we do not propose new mechanisms to account for the missing observations. Instead, we explore a novel aspect of the framework and shift the focus to checking whether it is possible to improve the performance of any instance of the framework by leveraging representation learning techniques. 

Having defined the \textsc{Rise} framework, we can formulate and reason about general modifications that can then be applied systematically to any instance of \textsc{Rise}. In particular, we explore the
following major changes:

\begin{enumerate}
    \item The conditionally replaced input $x^c_i$ is replaced with a \textit{latent} conditionally replaced input by mapping either the input signal (if observed) or the replacement input to a $d$-dimensional space via an appropriate encoding function: $\mathbf{e}^{x_i} = f^x_{\texttt{enc}}(x_i)$ and $\mathbf{e}^{\tilde{x}_i} = f^x_{\texttt{enc}}(\tilde{x}_i)$, respectively. 
    Therefore, the different parameters involved in the computation of the replacement input (see Table~\ref{instances:rise}) are adapted to this $d$-dimensional space. For instance, in \textsc{Rits-I} we replace the vector $\mathbf{w}_x$ with a matrix $\mathbf{W}_x$ of appropriate dimensions, and in \textsc{Gru-D} we learn latent representations for $x_{i'}$ and $x_{av}$ with the same encoder $f^x_{\texttt{enc}}$.
    \item For those instances that define a time gap vector, the time gap value $\delta_i$ is substituted with a $d_{\delta}$-dimensional latent time signal---for simplicity we set $d_{\delta}=d$---via an appropriate encoding function $\mathbf{e}^{\delta_i} = f^{\delta}_{\texttt{enc}}(\delta_i)$. This latent representation is to be used in the computation of the \textit{latent} discount factors $\mathbf{\gamma}^x \in \mathbb{R}^{d}$ and $\mathbf{\gamma}^h \in \mathbb{R}^{d_h}$.
\end{enumerate}

Appropriate encoders must be able to map single numerical values to a latent space. A dedicated encoding function is used for the input and time gap signal. Their parameters are learned jointly with all other parameters. We discuss a variety of encoding functions:

\xhdr{Feedforward-based encoders} A popular option \cite{li2017time,Pezeshkpour2018EmbeddingMR} for mapping single numerical values to a latent space is by applying a feedforward neural network over the (typically) log-normalized numerical value. Let $v$ and $\mathbf{e}^v$ be a numerical value and its corresponding latent representation, respectively, then this encoding function would amount to: 
\begin{equation}
\mathbf{e}^v \leftarrow f_{\texttt{ffw}}(v) = \sigma(\log(v) \mathbf{w} + b),
\end{equation}
where the numerical value $v$ only scales the weight vector $\mathbf{w} \in \mathbb{R}^{d}$, and $\sigma$ is a sigmoid activation function. It is easy to show that the learned latent representations behave monotonically (with respect to $v$) because of the monotonicity of the sigmoid function. To alleviate this limitation we explored a multi-feedforward encoding function with more layers, but surprisingly this deteriorated the performance.

We can overcome this limitation by applying a non-monotonic function such as the sine or the cosine. This is also inspired in part by Vaswani \textit{et al.}'s positional encoding \cite{vaswani2017attention}, which is extensively used in Transformer (\texttt{xfmr}) neural networks \cite{Devlin2019BERTPO}. For a given numerical value $v$, the representation learned by the encoder $f_{\texttt{xfmr}}$ is as follows:
\begin{equation}
\mathbf{e}^v \leftarrow f_{\texttt{xfmr}}(v) = [e^v_1;\dots; e^v_d] \quad
  \text{where }  e^v_k = \left\{
                \begin{array}{ll}
                  \sin(w_k v) \quad\text{if $k$ is even} \\
                  \cos(w_k v) \quad\text{if $k$ is odd}
                \end{array}
              \right.
\end{equation}
where $w_k$ is the frequency of the corresponding sinusoidal function. Vaswani \textit{et al.} \cite{vaswani2017attention} encoded discrete values---\ie positions of elements in a sequence---by fixing the wavelengths of the sinusoidal functions to form a geometric progression from $2\pi$ to $10,000 \cdot 2\pi$. Interestingly, they report similar performances in their downstream tasks when the frequencies are learned. Recent work by Kazemi \textit{et al.} \cite{Kazemi2019Time2VecLA} proposes a similar encoding function to learn representations for the time feature, and reports that learning the frequencies of the sinusoidal functions results in better performance. Hence we also allow the frequencies $\textbf{w}$ to be learned.

\xhdr{Binning encoder}
One may be tempted to transform numerical values to one-hot encoding representations, and learn latent representations via a linear transformation. However, the potential vocabulary size of this approach is infinite. Therefore a data binning scheme should be first applied to the data. In this work, we use quantiles, i.e., we define the width of the bins so as to ensure all $n_{\texttt{bin}}$ bins have a similar number of assigned data points.
Formally, a latent representation for the numerical value $v$ is obtained as 
\begin{equation}
\mathbf{e}^v \leftarrow f_{\texttt{bin}}(v) = \mathbf{W}_{\texttt{bin}} \texttt{one-hot}(\texttt{bin}(v)), 
\end{equation}
where $\mathbf{W}_{\texttt{bin}} \in \mathbb{R}^{d \times n_{\texttt{bin}}}$ is the bin embedding matrix and $\texttt{bin}$ is the chosen binning scheme. The downside of this approach is that the binning scheme is detached from the final learning objective.

\begin{table}[t]
\centering
\normalsize
\caption{\label{table:data} Statistics of the time series contained in the benchmark datasets.}
\resizebox{1\linewidth}{!}{
    \begin{tabular}{lrrr}
    \toprule
     \textbf{Dataset} &  \textbf{\# Observed Values  [$\%$]}     & \textbf{ \# Missing Values [$\%$]}         & \textbf{Range}      \\ \midrule
        BG@1&  372,404 [61$\%$]  & 243,639 [39$\%$]       & 40 - 400           \\ 
        BG@5&  106,975 [18$\%$]  &  509,068 [82$\%$]      & 40 - 400         \\ 
        Air Pollution - $\text{PM}_{10}$ & 173,243  [55$\%$]     & 142,117  [45$\%$]    & 5 - 1,000  \\ 
        \bottomrule
    \end{tabular}
    }
\vspace{-8pt}
\end{table}

\xhdr{Digit-level encoder} Motivated by character-level architectures for language modeling ~\cite{zhang2015character}, an alternative consists of decomposing numerical values into a sequence of digits and then operate on digits as atomic units to derive latent representations. The vocabulary consists of 11 tokens: the digits from 0 to 9, plus ``.'', which, if present, indicates the beginning of the decimal part of a numerical value. On some occasions, positive and negative signs may also be required in the vocabulary. We explore one approach to derive latent representations for numerical values from the embeddings of their digits: each token of the numerical value $v$ is mapped to its corresponding embedding of dimension $d_d$ via a linear transformation and the resulting sequence of embeddings is fed into a standard recurrent neural network architecture---a  GRU in this work. Let $n_v$ be the number of tokens of the numerical value $v$, its latent representation corresponds to the last hidden state $\mathbf{h}_{n_v} \in \mathbb{R}^{d}$ of the recurrent network: 

\begin{table*}[h!]
\fontsize{7}{7}\selectfont
  \caption{\label{results}Performance of different instances of the \textsc{Rise} framework with various encoding functions. The best result within each instance is always indicated in \textbf{bold}. MdAPE and MAPE is the median and mean APE, respectively.}
  \begin{subtable}[t]{0.4\linewidth}
\begin{tabular}{lcccccc}
\toprule
\multirow{2}[3]{*}{$f_{\texttt{enc}}$} & \multicolumn{2}{c}{BG@1} & \multicolumn{2}{c}{BG@5} & \multicolumn{2}{c}{AP - PM$_{10}$} \\
\cmidrule(lr){2-3} \cmidrule(lr){4-5} \cmidrule(lr){6-7}
 & MdAPE & MAPE & MdAPE & MAPE & MdAPE & MAPE \\
\midrule
$f_{\texttt{id}}$ &  2.3 & 4.3                & 6.8 & 12.0       & 14.0 & 26.6        \\ 
$f_{\texttt{ffw}}$ & 2.6 & 4.3                & 6.7 & 11.8         & 14.7 & 29.3    \\
$f_{\texttt{xfmr}}$ & \textbf{2.0} & 3.9                & \textbf{4.9} & \textbf{10.7}             & 14.8 & 27.1               \\
$f_{\texttt{bin}}$ & 2.3 & 4.1                & 5.7 & 10.8               & 14.1 & 26.4       \\
$f_{\texttt{gru}}$ & \textbf{2.0} &  \textbf{3.8}                & 5.2 & \textbf{10.7}               & \textbf{13.9} &  \textbf{26.0}   \\ 
\bottomrule
\end{tabular}
\caption{\textsc{Simple}}
  \end{subtable}
  \quad\quad\quad\quad\quad\quad\quad\quad\quad\quad
    \begin{subtable}[t]{0.4\linewidth}
\begin{tabular}{lcccccc}
\toprule
\multirow{2}[3]{*}{$f_{\texttt{enc}}$} & \multicolumn{2}{c}{BG@1} & \multicolumn{2}{c}{BG@5} & \multicolumn{2}{c}{AP - PM$_{10}$} \\
\cmidrule(lr){2-3} \cmidrule(lr){4-5} \cmidrule(lr){6-7} 
 & MdAPE & MAPE & MdAPE & MAPE & MdAPE & MAPE  \\
\midrule
$f_{\texttt{id}}$ &  2.5 & 4.3                & 6.3 & 11.5               & 14.3 & 27.9     \\ 
$f_{\texttt{ffw}}$ & 2.9 & 4.6                & 7.5 & 12.1               & 14.7 & 29.3      \\
$f_{\texttt{xfmr}}$ & \textbf{2.0} & \textbf{3.8}                & \textbf{5.1} & \textbf{10.4}                & 14.7 & 27.6            \\
$f_{\texttt{bin}}$ &  2.2 & 4.1                & 5.3 & 10.5               & 14.1 & \textbf{26.4}     \\
$f_{\texttt{gru}}$ & \textbf{2.0} & 3.9                & 5.2 & 10.5              & \textbf{13.8} & \textbf{26.4}  \\ 
\bottomrule
\end{tabular}
  \caption{\textsc{Zero-Filling\&Indicators}}
  \end{subtable}
  \\
    \begin{subtable}[t]{0.4\linewidth}
\begin{tabular}{lcccccc}
\toprule
\multirow{2}[3]{*}{$f_{\texttt{enc}}$} & \multicolumn{2}{c}{BG@1} & \multicolumn{2}{c}{BG@5} & \multicolumn{2}{c}{AP - PM$_{10}$} \\
\cmidrule(lr){2-3} \cmidrule(lr){4-5} \cmidrule(lr){6-7} 
 & MdAPE & MAPE & MdAPE & MAPE & MdAPE & MAPE \\
\midrule
$f_{\texttt{id}}$ &  2.2 & 4.1                & 6.0 & 11.1               & 14.0 & 26.5    \\
$f_{\texttt{ffw}}$ & 2.4 & 4.2                & 6.4 & 11.2               & 14.1 & 27.2    \\
$f_{\texttt{xfmr}}$ & \textbf{2.0} & 3.9                & 5.1 & 10.4               & 14.3 & 26.9  \\
$f_{\texttt{bin}}$ & 2.2 & 4.0                & 5.2 & 10.6               & 14.0 & 28.1      \\
$f_{\texttt{gru}}$ & \textbf{2.0} & \textbf{3.8}                & \textbf{5.0} &  \textbf{10.2}               & \textbf{13.8} & \textbf{25.5}    \\
\bottomrule
\end{tabular}
\caption{\textsc{Rits-I}}
  \end{subtable}
  \quad\quad\quad\quad\quad\quad\quad\quad\quad\quad
    \begin{subtable}[t]{0.4\linewidth}
\begin{tabular}{lcccccc}
\toprule
\multirow{2}[3]{*}{$f_{\texttt{enc}}$} & \multicolumn{2}{c}{BG@1} & \multicolumn{2}{c}{BG@5} & \multicolumn{2}{c}{AP - PM$_{10}$} \\
\cmidrule(lr){2-3} \cmidrule(lr){4-5} \cmidrule(lr){6-7}
 & MdAPE & MAPE & MdAPE & MAPE & MdAPE & MAPE  \\
\midrule
$f_{\texttt{id}}$ &  2.3 & 4.1                & 5.9 & 11.0               & 14.3 & 28.5       \\
$f_{\texttt{ffw}}$ & 2.5 & 4.2                & 6.6 & 11.2               & 15.4 & 28.0           \\
$f_{\texttt{xfmr}}$ & \textbf{2.0} & \textbf{3.8}                & 5.4 & 10.7               & 14.8 & 27.6   \\
$f_{\texttt{bin}}$ &  2.2 & 4.0                & 5.5 & 10.7               & \textbf{14.2} & 28.1    \\
$f_{\texttt{gru}}$ & \textbf{2.0} & \textbf{3.8}                & \textbf{5.1} & \textbf{10.4}               & \textbf{14.2} & \textbf{26.5}      \\ 
\bottomrule
\end{tabular}
  \caption{\textsc{GRU-D}}
  \end{subtable}
  \vspace{-15pt}
\end{table*}

\begin{equation}
\mathbf{e}^v \leftarrow f_{\texttt{gru}}(v) =\mathbf{h}_{n_v}. 
\end{equation}

Contrary to previous encoding functions, digit-level encoders are neither monotonic nor detached from the final learning objective. However, their learned representations are driven---and limited---by the atomic units the numerical values are decomposed into. Contrary to language, similarities across numerical values based on its atomic units is arbitrary and based on the chosen notation scheme.

\section{Experiments}
\label{sec:exps}
We focus on the following instances of the framework: Simple Recursion (\textsc{Simple}), Zero-\textsc{Filling$\&$Indicators}, \textsc{Rits-I} and \textsc{Gru-D}. We choose these ones as most of the other instances listed in Table \ref{instances:rise} are constituted by combinations of these approaches. We equip these instances with the encoding functions described in Section \ref{sec:repr}. As these \textsc{Rise} instances work in a forward manner, we simply pretend the next observed value of the time series is missing and aim to impute it. Models are evaluated at any point in time in which there are at least ten samples of prior data, and the accuracy is defined in terms of the absolute percentage error (APE) \cite{fox2018deep}. The APE between a prediction $\hat{y}$ and a ground truth value $y$ is defined as follows: $\text{APE}(y_i, \hat{y}_i) = 100 \times \left| \cfrac{y_i - \hat{y}_i}{y_i} \right|$. We report median and mean APE for all experiments. To ensure a fair comparison, we implemented all the instances of the \textsc{Rise} framework and ran experiments under exactly the same setup and evaluation protocol. 

\xhdr{Datasets} We use the blood glucose dataset used in previous work \cite{fox2018deep}, which consists of a large number of continuous glucose readings from 40 patients with type 1 diabetes. We run experiments in two versions of the dataset, denoted BG@1 and BG@5, which contain different proportions of missing observations. We use the same training, validation and test sets as in previous work \cite{fox2018deep}. We also use an air pollution dataset \cite{Yi2016STMVLFM}, which consists of measurements of $\text{PM}_{10}$ from air monitoring stations in Beijing. The first ten months of data are used for training, month 11th for validation, and month 12th for testing. 
Dataset statistics are shown in Table \ref{table:data}. They are available at \url{https://doi.org/10.5281/zenodo.4117595}.\\



\xhdr{Setup} We use the same evaluation protocol and setup as Fox \textit{et al.} \cite{fox2018deep}: all instances are built by stacking two GRUs with $d_h=512$ hidden units in a standard manner \cite{hermans2013training}, and trained with the multi-classification formulation of the framework (see Section \ref{sec:rise}). Performance is validated according to both the median and mean APE on the validation set after every epoch, and store the best validated model for each metric. We apply $L_2$ regularization to the parameters of the models and validate the regularization term among $\{10^{-2}, 10^{-3}, 10^{-4}\}$. Models were trained for 100 epochs, and used the Adam \cite{kingma:adam} optimizer. The dimension $d$ of the different encoding functions is validated between $\{64, 128\}$. The dimension $d_d$ of the digit-level encoder is fixed to 64. The number of bins of the binning encoder is validated among the values $\{10, 50, 100\}$.

\xhdr{Results}
Results in the test sets are depicted in Table \ref{results}. The identity encoder simply returns the input unmodified---$f_{\texttt{id}}(v) =v$. That is, the identity encoders correspond to the \textsc{Rise} instances as defined in their respective papers.



\noindent\textit{Impact of $\tilde{x}$, $h_x$ and $g_h$ (when encoder is $f_{\texttt{id}}$):} The similarities, highlighted in Table \ref{instances:rise}, between the instances \textsc{Simple} and \textsc{Rits-I} translate into similar performance, with \textsc{Rits-I} showing a slight superior performance. We hypothesize that the main difference in performance between both methods, observed in BG@5, is caused by the exponential discount factor of \textsc{Rits-I} applied to the hidden state, as large blocks of observations are missing in this time series. The only difference between \textsc{Rits-I} and \textsc{Gru-D} is in the computation of the replacement input $\tilde{x}$. However, while they perform similarly in the blood glucose data, they differ in terms of mean APE in the air pollution time series. This showcases that the importance of the different modules of the framework ($\tilde{x}$, $h_x$ and $g_h$) may vary across datasets. Overall,  \textsc{Zero-Filling\&Indicators} tends to show a less competitive performance in most of the datasets. The results  shows that i) there is not a clear instance of the framework that systematically performs best; ii) simple recursion (\textsc{Simple}) seems to be a reasonable solution in almost all datasets.
\begin{figure}
\centering
   \includegraphics[width=0.55\linewidth]{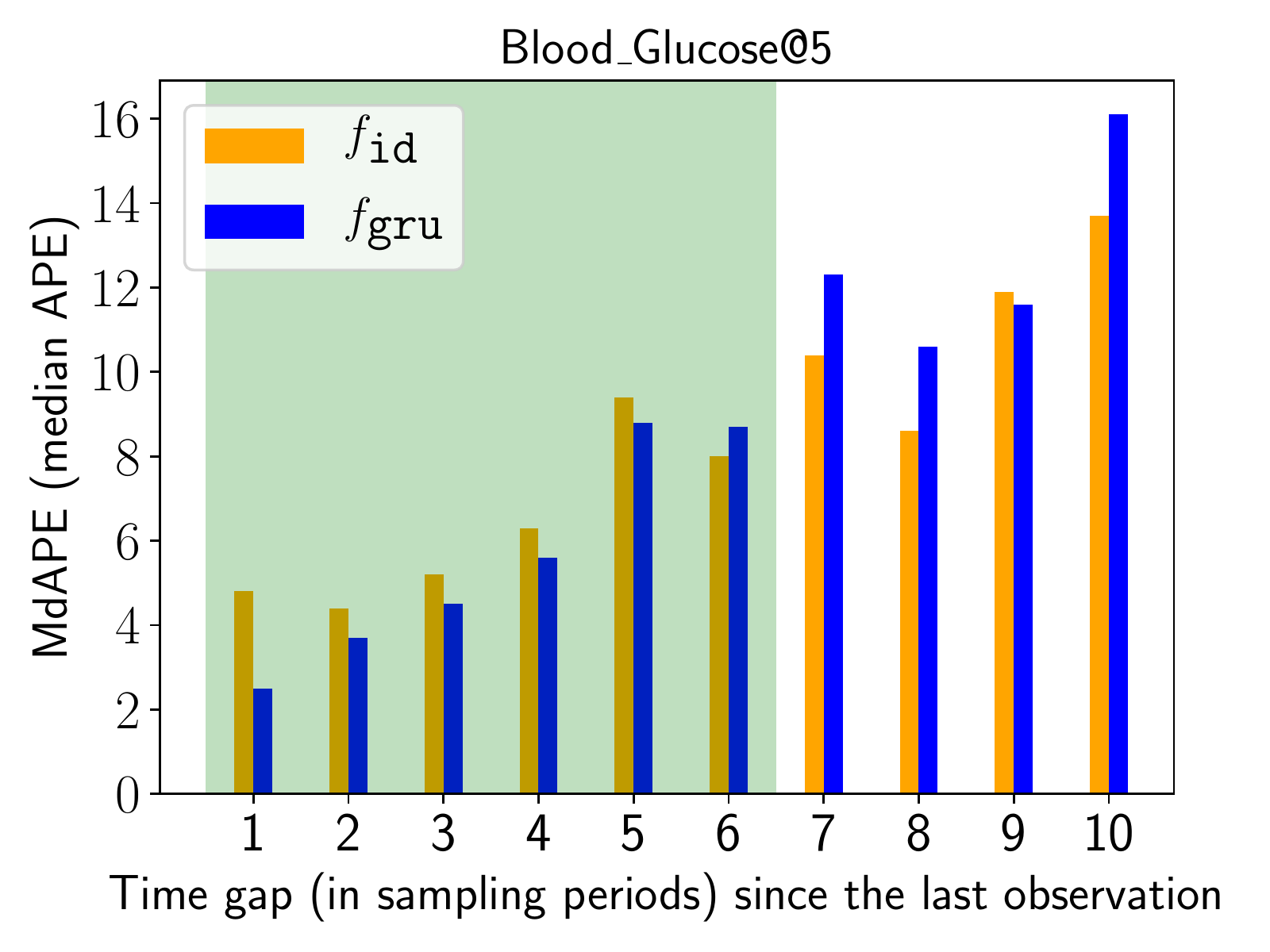}
\caption{\label{perf_lag} Performance breakdown $f_{\texttt{gru}}$ \textit{vs} $f_{\texttt{id}}$}
\vspace{-15pt}
\end{figure}

\noindent\textit{Impact of the encoder $f_{\texttt{enc}}$:} Although it has been used in other problems~\cite{li2017time,Pezeshkpour2018EmbeddingMR}, we observe that the feedforward encoding function $f_{\texttt{ffw}}$ deteriorates the performance of the evaluated instances in this problem. On the contrary, the digit-level encoders show improvements in both metrics for every instance of the framework and every dataset. As discussed in Section \ref{sec:repr}, the representations learned by the digit-level encoders are meaningful with respect to the learning objective. Such homogeneous benefit is not observed in the binning encoder. As also discussed in Section \ref{sec:repr}, we hypothesize that this is because the binning strategy is detached from the learning objective. A different choice of binning scheme might also lead to improvements in the air pollution dataset, but the search for an appropriate binning scheme may be very costly. Similar to the binning encoder, the feedforward encoder with sinusoidal activation functions $f_{\texttt{xfmr}}$ is not consistent across all datasets. 

Figure \ref{perf_lag} breaks down the median APE of the predictions by \textsc{Simple} in BG@5 into the number of sampling periods since the last observation. Previous work \cite{Turksoy2013HypoglycemiaEA,Plis2014AML,fox2018deep} has settled on a 30-minute prediction window as adequate for a similar task and dataset. This time window is represented with a green background in Figure \ref{perf_lag}. Our experiments show that the improvements of $f_{\texttt{gru}}$ over $f_{\texttt{id}}$ are mainly gained in this time window.

\section{Conclusion}
\label{sec: Conclusion}
We introduce \textsc{Rise}, a unifying framework that encompasses as special cases multiple recent approaches for learning from time series with missing data. We showcase this conceptual strength by proposing novel input and state encoding functions and plugging them into multiple previously proposed methods that can be seen as instances of \textsc{Rise}. Our evaluation on three data imputation tasks shows that a digit-level encoder always performs best. Future work should adapt this encoder to work with multivariate time series. 

\bibliographystyle{IEEEbib}
\bibliography{main}

\end{document}